\documentclass{article}

\usepackage[preprint]{neurips_2026}
\workshoptitle{Dynamic Alignment on Human-AI Coupled System (BiAlign Workshop)}
\usepackage[utf8]{inputenc}
\usepackage[T1]{fontenc}
\usepackage{courier}
\usepackage{amsmath,amssymb,mathtools}
\usepackage{booktabs}
\usepackage{graphicx}
\usepackage{float}
\usepackage{microtype}
\usepackage{xcolor}
\usepackage{hyperref}
\usepackage{url}
\usepackage{tabularx}
\usepackage{array}
\usepackage{enumitem}
\usepackage{tikz}
\usetikzlibrary{arrows.meta,positioning,fit,calc}
\setcitestyle{numbers,square}

\definecolor{lqetblue}{HTML}{315B7D}
\definecolor{taskorange}{HTML}{C45A23}
\definecolor{softgray}{HTML}{F2F4F6}

\hypersetup{
  colorlinks=true,
  linkcolor=lqetblue,
  citecolor=black,
  urlcolor=lqetblue
}

\newcommand{\method}{\textsc{LQET}}
\newcommand{\yes}{\texttt{YES}}
\newcommand{\no}{\texttt{NO}}
\newcommand{\KL}{D_{\mathrm{KL}}}
\newcommand{\modelscore}{z_{\theta}}
\newcommand{\targetprob}{p^{\star}}

\title{Provenance-Aware Belief Updating\\
in Human-AI Feedback Loops}

\author{%
  Tianyi Huang \\
  Ryquo
  \And
  Eric Liu \\
  App-In Club
  \And
  Jay Chong \\
  App-In Club
  \And
  Emily Yin \\
  App-In Club
  \And
  Wanqi Zhao \\
  App-In Club
  \AND
  \normalfont
  \texttt{tianyi@ryquo.com}\quad
  \texttt{\{eric.liu,jay.chong,emily,vicky\}@appinclub.org}
}
\begin{document}

\maketitle

\begin{abstract}
In repeated human-AI interaction, a single observation can be quoted, summarized, and returned to a model as multiple reports, making repetition look like independent corroboration. We study whether language models can use supplied report lineage to update according to evidence provenance rather than report count. We introduce Latent Quotient Evidence Training (\method), a relational objective that combines sparse Bayesian supervision with constraints that encourage stable predictions when reports add no new evidence and consistent updates to the same new evidence across relay histories. On a controlled binary benchmark, \method{} reduces new-evidence update RMSE by 41--44\% and log-odds drift after eight relay steps by 89--91\% relative to task-only training across Qwen3-4B and Phi-4-mini, with three paired training seeds per backbone. The gains extend to a closed-loop simulation in which model confidence controls how many relayed reports re-enter the model's context in later rounds. By grounding belief updates in supplied provenance, \method{} offers a route toward dynamic human-AI systems in which influence follows new evidence, not the number of times the same evidence returns.
\end{abstract}

\section{Introduction}
\label{sec:intro}

\begin{figure}[t]
  \centering
  \includegraphics[width=\linewidth]{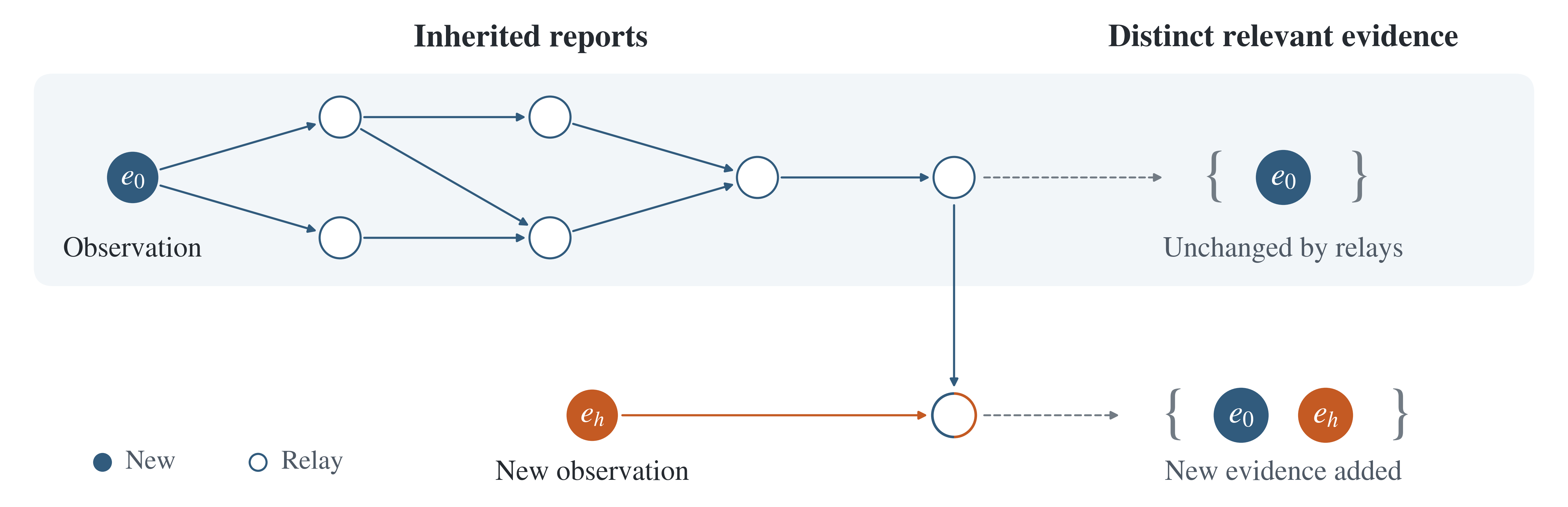}
  \caption{\textbf{Reports can multiply without adding evidence.}
  Relays preserve the evidence set $\{e_0\}$; a new observation expands it to
  $\{e_0,e_h\}$. Solid arrows show report derivation; dashed arrows identify
  distinct evidence. With the prior and calibration fixed, histories with the
  same evidence share a Bayesian target.}
  \label{fig:provenance-overview}
\end{figure}

A report is not always a new observation. In a coupled information system, one claim may be quoted by a model, copied into a summary, and returned through later context. A single speaker can sound like a chorus even when listeners know that repeated statements share a source \citep{weaver2007chorus}. In Bayesian social learning and decentralized estimation, inadvertent reuse of the same information (data incest) can bias beliefs and produce overconfidence \citep{brehard2007optimal,krishnamurthy2013misinformation}. People also tend to neglect correlation among signals \citep{enke2019correlation}, while repeated human-AI interaction can amplify judgment biases through feedback \citep{glickman2025feedback}. Appropriate updating therefore depends not only on what reports say, but also on whether they introduce evidence or inherit it (Figure~\ref{fig:provenance-overview}).

Prior work examines probabilistic updating from demonstrations
\citep{qiu2026bayesian}, responses to evidence with varying informativeness and reliability \citep{kim-etal-2025-evidence}, sequential belief trajectories \citep{samanta2026bayesbench}, and sensitivity to repeated or paraphrased support \citep{schuster-etal-2026-whose, naphade-2026-rational}. We ask a complementary structured question: given explicit multi-hop report provenance, can a model preserve its prediction when a report merely relays existing evidence, yet apply the same reliability-conditioned update when genuinely new evidence arrives after different relay histories? Sparse pointwise supervision on individual transcript views does not directly enforce this cross-transcript relationship: it assigns each transcript its own target, whereas the desired behavior concerns equivalence classes of transcripts and score changes between them.

We formalize this supplied-lineage setting with matched families of synthetic evidence reports. Prompts expose the derivation graph through parent-report links and natural-language descriptions of whether a report introduces a distinct observation or relays an earlier one; evidence-root identifiers used to construct the matched families are not rendered. Two transcripts are provenance-equivalent when they contain the same set of distinct, query-relevant evidence roots. Under the source-reliability model, exact Bayesian inference assigns the same log-odds target to every transcript in an equivalence class. The model should therefore satisfy two relations: adding only reports derived from existing evidence should leave its score unchanged, whereas adding a new evidence root should produce the same score change regardless of the preceding relay history.

Latent Quotient Evidence Training (\method) directly supervises these relations. It combines sparse soft-target supervision with four invariance constraints, three new-evidence update-consistency constraints, and a presentation-order constraint. This isolates a prerequisite for provenance-aware systems: using available lineage consistently when inherited reports return, with provenance discovery treated as a separate problem.

Our contributions are:

\begin{itemize}[leftmargin=*,itemsep=2pt,topsep=3pt]
    \item We formalize provenance equivalence over sets of distinct evidence roots and derive exact Bayesian log-odds targets under uncertain source reliability.
    \item We introduce a relational objective that extends repetition invariance with new-evidence update consistency, composition, and presentation-order consistency while retaining sparse pointwise supervision.
    \item Across two language-model backbones, paired training seeds, and controlled distribution shifts, \method{} consistently improves new-evidence update accuracy and stability across long relay histories.
    \item A confidence-dependent feedback simulation connects these static relations to the dynamic setting in which model confidence influences how many inherited reports return as later input.
\end{itemize}

\section{Evidence provenance and Bayesian targets}
\label{sec:setup}

\paragraph{Latent state and source reliability.}
Let $Y\in\{-1,+1\}$ be the binary answer with prior $\pi=P(Y=+1)\in(0,1)$. Independently of $Y$, the source reliabilities $R_s$ have mutually independent $\operatorname{Beta}(1,1)$ priors before observing calibration data. Conditional on $R_s$, the source's calibration outcomes are independent and identically distributed Bernoulli trials; a visible record of $c_s$ correct and $w_s$ incorrect measurements therefore gives
\begin{equation}
R_s\mid c_s,w_s\sim\operatorname{Beta}(\alpha_s,\beta_s),
\qquad \alpha_s=c_s+1,\quad \beta_s=w_s+1.
\label{eq:reliability}
\end{equation}
An evidence root $e$ denotes one distinct underlying observation, with reported outcome $x_e\in\{-1,+1\}$ and source $s(e)$. For a query-relevant root,
$P(x_e=Y\mid Y,R_{s(e)})=R_{s(e)}$. Conditional on $Y$ and $\{R_s\}_s$, the outcomes of distinct roots are independent of one another and of the calibration outcomes. Multiple roots from the same source share $R_s$ and thus become dependent after reliability is marginalized.
By construction, conditional on the relevant-root observations and visible calibration records, the report graph, rendering choices, and query-irrelevant roots carry no additional information about $Y$.

\paragraph{Reports and provenance equivalence.}
A report is a node in an acyclic derivation graph. It may link to parent reports, inherit their evidence roots, and introduce zero or more roots of its own. For transcript $T$, let
\begin{equation}
E(T)=\bigcup_{r\in T}\{e:e\text{ is a query-relevant root represented by }r\}
\label{eq:quotient}
\end{equation}
be the set of \emph{distinct} relevant roots. Set union counts each root once, regardless of how many reports inherit it. Within a matched family, we define
\begin{equation}
T\sim T'\quad\Longleftrightarrow\quad E(T)=E(T').
\end{equation}
This relation partitions each matched family into provenance-equivalence classes and induces the quotient used by \method{}. To compute the Bayesian target, let
$n_{s,a}(T)=|\{e\in E(T):s(e)=s,\ x_e=a\}|$ for $a\in\{-1,+1\}$.

\paragraph{Exact Bayesian target.}
For $y\in\{-1,+1\}$, let
\begin{equation}
(C_s(y,T),W_s(y,T))=
\begin{cases}
(n_{s,+}(T),n_{s,-}(T)) & y=+1,\\
(n_{s,-}(T),n_{s,+}(T)) & y=-1.
\end{cases}
\end{equation}
Let $B(a,b)=\Gamma(a)\Gamma(b)/\Gamma(a+b)$ and $\sigma(u)=(1+e^{-u})^{-1}$. All logarithms are natural.
Write $\mathcal O(T)=\bigl((s(e),x_e)\bigr)_{e\in E(T)}$ for the indexed collection of relevant-root observations. After conditioning on the calibration records, let $\ell_y(T)$ be the log prior plus log marginal likelihood for $Y=y$; it equals the posterior log-probability up to an additive normalizing constant that is independent of $y$. Integrating each $R_s$ yields
\begin{align}
\ell_y(T)
&=\log P(Y=y)
+\sum_s\left[
\log B(\alpha_s+C_s(y,T),\beta_s+W_s(y,T))
-\log B(\alpha_s,\beta_s)
\right],\\
t(T)&=\ell_{+1}(T)-\ell_{-1}(T),
\qquad \targetprob(T)=P\!\left(Y=+1\mid\mathcal O(T),\{c_s,w_s\}_s\right)=\sigma(t(T)).
\label{eq:bayes-target}
\end{align}
Thus a relay that adds no new evidence leaves $t(T)$ unchanged. The update from a new root depends on its source, reported outcome, and the other roots from that source. The Bayesian target is constant within each provenance-equivalence class and, holding the prior and calibration records fixed, depends on a root set only through its source-by-outcome counts under the exchangeable source model.

\paragraph{Matched transcript families.}
Each static benchmark family produces 14 matched transcript views. The core $2\times2$ block crosses a relay history with the addition of new evidence: $h_{00}$ contains the base root, $h_{10}$ adds only relays, $h_{01}$ adds a new root, and $h_{11}$ adds that root after the relay history. Hence
\begin{equation}
t(h_{10})=t(h_{00}),\qquad t(h_{11})=t(h_{01}),\qquad
t(h_{11})-t(h_{10})=t(h_{01})-t(h_{00}).
\label{eq:core-identities}
\end{equation}
A composition block sharing $c_{00}=h_{00}$ separately adds roots $e_a$ and $e_b$ in $c_{10}$ and $c_{01}$, adds both in $c_{11}$, and reverses their presentation in $c_{11}^{\mathrm{rev}}$. The base root and both added roots are drawn from three mutually distinct sources. Matched controls hold report author and displayed outcome fixed while varying the supplied provenance: a report either relays an existing observation or introduces a new one. Prompts include priors, calibration records, source names, report text, and explicit parent-report links; internal evidence-root and family identifiers are never rendered. A deterministic prompt-only calculator recovers every static-view Bayesian
target, confirming that each target is fully specified by the visible prompt (Appendix~\ref{app:identifiability}).

\begin{figure}[t]
  \centering
  \includegraphics[width=\linewidth]{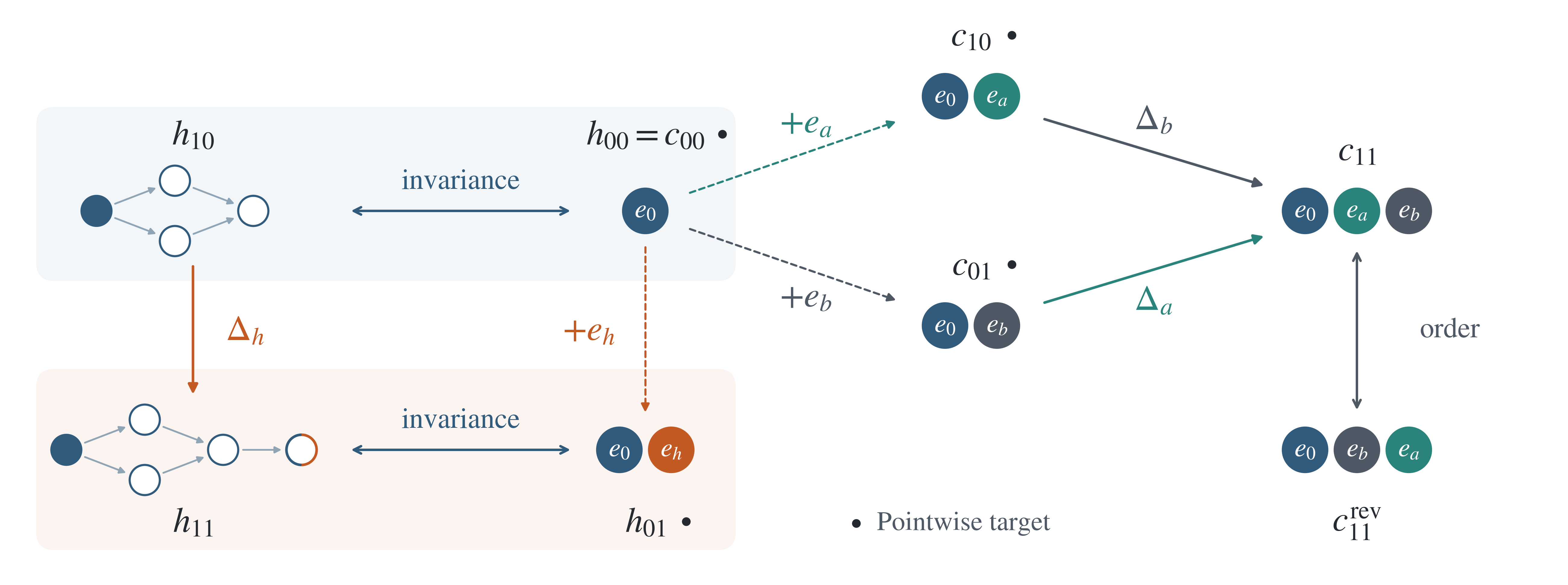}
  \caption{\textbf{Selected LQET training relations.}
  LQET trains matched transcripts to preserve scores under relays and reordering,
  and to match Bayesian updates when new evidence is added. Shading groups
  equal-evidence views; dots mark pointwise targets.
  $\Delta_h,\Delta_a,\Delta_b$ denote target updates for adding
  $e_h,e_a,e_b$, respectively. The composition example uses three distinct sources.}
  \label{fig:lqet-relations}
\end{figure}

\section{Latent Quotient Evidence Training}
\label{sec:method}

\paragraph{Forced-choice score.}
We evaluate a next-token forced-choice diagnostic rather than parse a generated numeric response. For prompt $T$, the score is the \yes-\no{} log-odds
\begin{equation}
\modelscore(T)=L_\theta(\yes\mid T)-L_\theta(\no\mid T),
\qquad \hat p_\theta(T)=\sigma(\modelscore(T)).
\label{eq:model-score}
\end{equation}
Here $L_\theta(a\mid T)$ is the pre-softmax next-token logit for token $a$. Thus $\hat p_\theta(T)=P_\theta(\yes\mid T,\text{next token}\in\{\yes,\no\})$: it is a forced-choice \yes{} score, not the full-vocabulary probability of an unconstrained response. Each tokenizer maps \yes{} and \no{} to distinct single tokens at the assistant boundary.

\paragraph{Sparse pointwise supervision.}
Seven views receive pointwise soft targets. Writing
\[
\mathcal A=\{h_{00},h_{01},c_{10},c_{01},\text{same-source},\text{counter},\text{matched-new}\},
\]
let $N$ be the number of matched families in a training batch and define $\targetprob_{i,v}=\targetprob(T_{i,v})$ and $\hat p_{i,v}=\hat p_\theta(T_{i,v})$. Their average Bernoulli cross-entropy is
\begin{equation}
\mathcal L_{\mathrm{task}}
=-\frac{1}{N|\mathcal A|}\sum_{i=1}^N\sum_{v\in\mathcal A}
\left[\targetprob_{i,v}\log\hat p_{i,v}+(1-\targetprob_{i,v})\log(1-\hat p_{i,v})\right].
\label{eq:task}
\end{equation}
Views such as $h_{11}$ and $c_{11}$ in Figure~\ref{fig:lqet-relations} are supervised only through relational losses and receive no pointwise loss.

\paragraph{Relational constraints.}
For family $i$, write $z_i(v)=z_\theta(T_{i,v})$ and $t_i(v)=t(T_{i,v})$. We use the Huber penalty with $\delta=0.25$,
\begin{equation}
\rho_\delta(u)=
\begin{cases}
\tfrac12 u^2, & |u|\leq\delta,\\
\delta\bigl(|u|-\tfrac12\delta\bigr), & |u|>\delta.
\end{cases}
\label{eq:huber}
\end{equation}
The four provenance-invariance residuals are
\begin{align*}
r^{\mathrm{inv}}_{i,1}&=z_i(h_{10})-z_i(h_{00}), &
r^{\mathrm{inv}}_{i,2}&=z_i(h_{11})-z_i(h_{01}),\\
r^{\mathrm{inv}}_{i,3}&=z_i(\text{matched-copy})-z_i(h_{00}), &
r^{\mathrm{inv}}_{i,4}&=z_i(\text{irrelevant})-z_i(h_{00}),
\end{align*}
and their mean loss is
\begin{equation}
\mathcal L_{\mathrm{inv}}
=\frac{1}{4N}\sum_{i=1}^{N}\sum_{k=1}^{4}\rho_\delta\!\left(r^{\mathrm{inv}}_{i,k}\right).
\label{eq:invariance}
\end{equation}
The three new-evidence update residuals are
\begin{align*}
r^{\mathrm{upd}}_{i,1}
&=[z_i(h_{11})-z_i(h_{10})]-[t_i(h_{01})-t_i(h_{00})],\\
r^{\mathrm{upd}}_{i,2}
&=[z_i(c_{11})-z_i(c_{10})]-[t_i(c_{01})-t_i(c_{00})],\\
r^{\mathrm{upd}}_{i,3}
&=[z_i(c_{11})-z_i(c_{01})]-[t_i(c_{10})-t_i(c_{00})],
\end{align*}
with
\begin{equation}
\mathcal L_{\mathrm{update}}
=\frac{1}{3N}\sum_{i=1}^{N}\sum_{k=1}^{3}\rho_\delta\!\left(r^{\mathrm{upd}}_{i,k}\right).
\label{eq:update}
\end{equation}
The target differences in the latter two residuals use the benchmark's three-way source distinctness: the base root and the two added roots come from mutually distinct sources. Updates from multiple roots sharing a source need not be additive. Finally,
$\mathcal L_{\mathrm{order}}=N^{-1}\sum_i\rho_\delta(z_i(c_{11})-z_i(c_{11}^{\mathrm{rev}}))$.
The implemented objective is
\begin{equation}
\boxed{\mathcal L_{\method}=\mathcal L_{\mathrm{task}}+4\mathcal L_{\mathrm{inv}}+2\mathcal L_{\mathrm{update}}+\mathcal L_{\mathrm{order}}.}
\label{eq:lqet}
\end{equation}
Each component is averaged before weighting, giving per-residual coefficients of $1$, $2/3$, and $1$ for invariance, update, and order, respectively, within a family. Appendix~\ref{app:weight-sensitivity} explains this scaling and the available sensitivity evidence. \method{} uses 13 training views versus seven for task-only, with identical families and optimizer-update counts.

\section{Experimental design}
\label{sec:experiments}

\paragraph{Backbones and training.}
We evaluate Qwen3-4B-Instruct-2507 \citep{qwen2025qwen3,qwen2025qwen3_4b_instruct_2507} and Phi-4-mini-instruct \citep{microsoft2025phi4mini}. For each backbone and method, training seeds 0, 1, and 2 use the same fixed families: 4,000 for training and 400 for validation. We train for 500 updates (two epochs) with QLoRA \citep{hu2022lora,dettmers2023qlora}: NF4 with double quantization, BF16 computation, LoRA rank 32, scaling factor $\alpha=64$, dropout 0.05, and adapters on all linear modules. AdamW uses learning rate $2\times10^{-4}$, zero weight decay, a 3\% warmup followed by cosine decay, gradient accumulation 16, and clipping at 1.0. Sequence length is capped at 768, and no evaluated prompt is truncated. The $1{:}4{:}2{:}1$ loss-weight ratio, Huber threshold, and optimization settings were fixed before the reported Qwen runs and transferred unchanged to Phi. Validation results were used only for run diagnostics, without hyperparameter or checkpoint selection. We evaluate the final checkpoint.

\paragraph{Evaluation sets.}
Each run is evaluated on 400 independent and identically distributed (IID) test families from the training distribution, 800 controlled-shift out-of-distribution (OOD) families (160 for each of five factor groups: held-out domains and source names, longer relay histories, fan-in and diamond provenance graphs, held-out templates, and held-out prior and calibration-count supports), and 200 recursive-relay families at depths $1,2,4,8$. All 14 views are scored for the IID and OOD families; the recursive-relay families are evaluated on nested relay histories and their subsequent new-evidence probes. For the static IID and OOD views, task-only and \method{} are scored on identical rendered prompts. Recursive-relay prompts are model-specific because each appended relay quotes that model's preceding forced-choice score, although the underlying family, lineage schedule, and rendering rule are aligned. All partitions are mutually disjoint, and we check for overlap in sampled benchmark instances, evidence roots, family identifiers, and rendered prompts.

\paragraph{Primary endpoints.}
For the primary Qwen evaluation and Phi replication, new-evidence update RMSE and supervised-view KL pool the 800 families across the five equally sized OOD shifts within each seed; shift-specific estimates are secondary. New-evidence update RMSE compares predicted log-odds changes with the corresponding exact Bayesian changes on the three update pairs in Equation~\ref{eq:update}. Depth-8 relay drift averages the absolute score change from the base transcript after eight recursively generated relays over the 200 relay families. Supervised-view KL compares the exact Bayesian target with the forced-choice Bernoulli score and averages $\KL(\operatorname{Bern}(\targetprob)\Vert\operatorname{Bern}(\hat p))$ over the seven pointwise-supervised views. Each backbone-level point estimate is the arithmetic mean of its three seed-level estimates; for RMSE, the square root is taken within each seed before averaging.

\paragraph{Statistical inference and decision criterion.}
We use 10,000 paired crossed seed-by-family bootstrap replicates. For each endpoint, a replicate independently resamples the three paired training seeds and one shared evaluation-family roster; the same family resample is reused across every selected seed occurrence and both methods, with all views from a family kept together. Pooled OOD resampling is stratified over the five fixed 160-family shifts, and nonlinear endpoints are recomputed within seed before averaging across resampled seeds. The 2.5th and 97.5th percentiles define unadjusted 95\% intervals (Appendix~\ref{app:protocol}). The prespecified conjunctive criterion requires (i) negative \method{}-minus-task-only effects for update RMSE and depth-8 relay drift in every seed, with pooled 95\% upper bounds below zero, and (ii) an \method{}-minus-task-only supervised-view KL effect of at most 0.02 in every seed, with a pooled upper bound of at most 0.02. The $+0.02$-nat non-inferiority margin for mean pointwise prediction quality was fixed before training. We analyze each backbone separately and do not form a cross-backbone interval.

\section{Results}
\label{sec:results}

\begin{figure}[t]
    \centering
    \includegraphics[width=\linewidth]{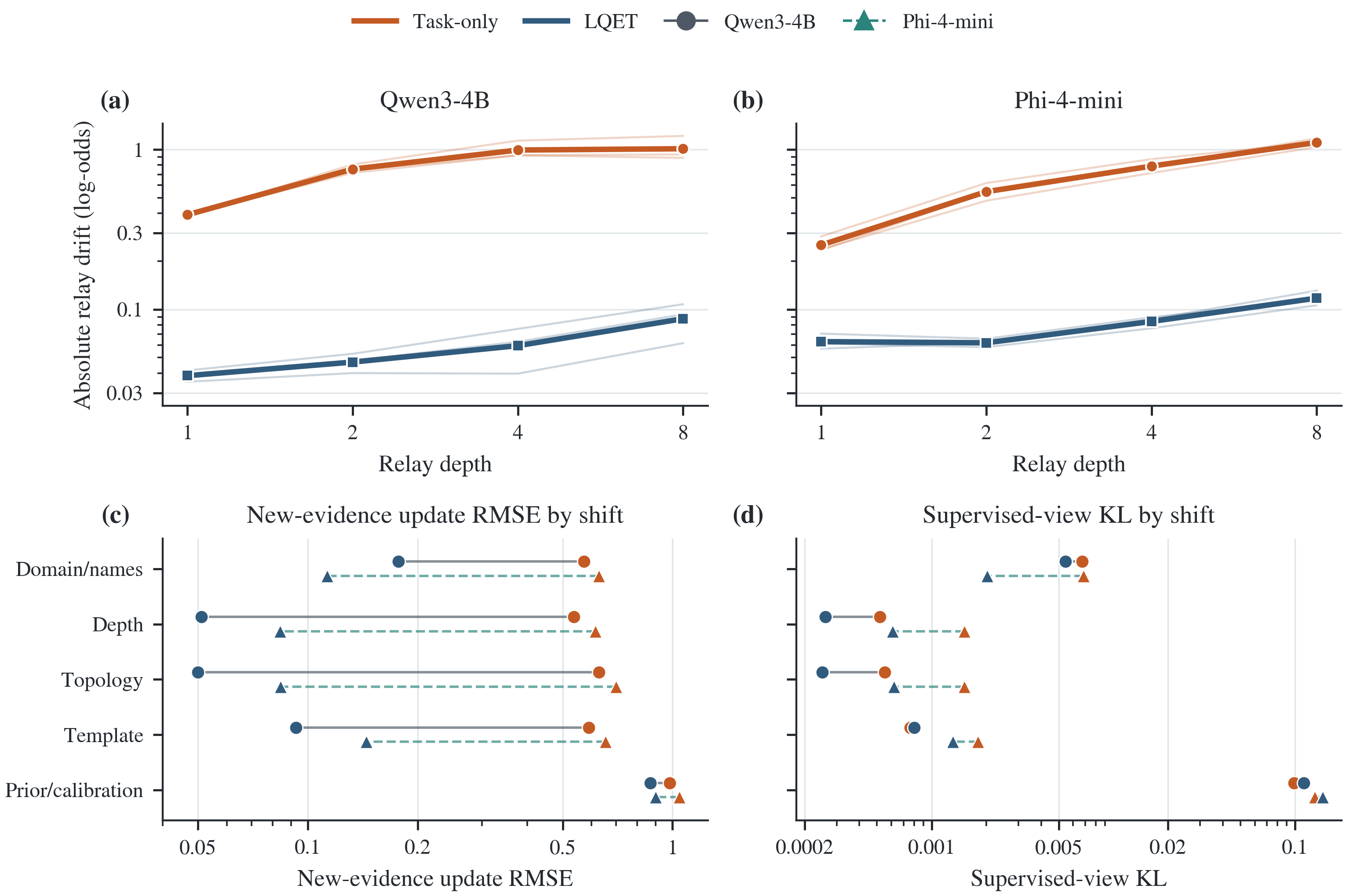}
    \caption{\textbf{Controlled generalization across relay depth and distribution shift.} (a,b) Absolute relay-induced drift across nested depths; thick and thin curves show three-seed means and individual seeds on the same 200 families. (c) Update RMSE and (d) supervised-view KL across controlled shifts; color marks the method, shape the backbone, and connectors the paired contrast. Panels (a,b) use logarithmic y-axes, and panels (c,d) use logarithmic x-axes. Lower is better.}
    \label{fig:controlled-generalization}
\end{figure}

\subsection{Controlled results across backbones}

\begin{table}[h]
\centering
\caption{\textbf{Primary controlled results.} Effects are \method{} minus task-only and are computed from unrounded estimates; lower is better. Point estimates average three paired training seeds. Intervals are 95\% paired crossed seed-by-family bootstrap intervals. $\dagger$ marks the supervised-view KL endpoint with a prespecified $+0.02$-nat non-inferiority margin.}
\label{tab:primary}
\small
\setlength{\tabcolsep}{4.3pt}
\begin{tabular}{llrrr}
\toprule
Backbone & Endpoint & Task-only & \method{} & Effect [95\% interval] \\
\midrule
Qwen3-4B & Update RMSE & 0.6816 & \textbf{0.4005} & $-0.2812$ [$-0.3070,-0.2565$] \\
& Depth-8 drift & 1.0122 & \textbf{0.0875} & $-0.9247$ [$-1.1175,-0.7703$] \\
& Supervised-view KL$^{\dagger}$ & 0.0215 & 0.0237 & $+0.0023$ [$+0.00116,+0.00356$] \\
\addlinespace
Phi-4-mini & Update RMSE & 0.7461 & \textbf{0.4144} & $-0.3317$ [$-0.3545,-0.3091$] \\
& Depth-8 drift & 1.1079 & \textbf{0.1177} & $-0.9902$ [$-1.0823,-0.8988$] \\
& Supervised-view KL$^{\dagger}$ & 0.0280 & 0.0292 & $+0.0012$ [$-0.00002,+0.00251$] \\
\bottomrule
\end{tabular}
\end{table}

Table~\ref{tab:primary} shows the same pattern on both backbones. \method{} lowers new-evidence update RMSE by 41.25\% on Qwen and 44.45\% on Phi. Depth-8 relay drift falls by 91.36\% and 89.38\%, respectively. Both improvements hold in every training seed, and the three-seed mean update RMSE is lower on every OOD shift for both backbones. Pooled supervised-view KL changes by $+0.00228$ on Qwen and $+0.00120$ on Phi, with both upper bootstrap bounds below the prespecified $+0.02$-nat non-inferiority margin. Both backbones therefore satisfy the full conjunctive criterion. On the IID test set, \method{} lowers both update RMSE and supervised-view KL on both backbones (Appendix~\ref{app:primary-details}, Table~\ref{tab:iid-results}).

Figure~\ref{fig:controlled-generalization} breaks down the aggregate results by relay depth and distribution shift. Along nested relay histories, task-only drift accumulates sharply through depth four and remains near one log-odds unit at depth eight, whereas the three-seed mean for \method{} remains below 0.14 at every depth (panels a,b). Update RMSE is lower on every controlled shift (panel c), with the largest proportional gains on unseen depth and topology and smaller gains on held-out prior and calibration supports. Supervised-view KL is lower or essentially unchanged on the other four shifts but increases on the held-out prior-and-calibration shift, the only shift that changes the numerical supports for both the class prior and source-calibration records (panel d). Each recursive relay quotes the preceding forced-choice \yes{} score rounded to three decimals, so the depth test repeatedly introduces model-generated surface content rather than duplicating a fixed string.

\subsection{Selective invariance, not uniform update suppression}
\label{sec:mechanism-results}

\begin{figure}[t]
    \centering
    \includegraphics[width=\linewidth]{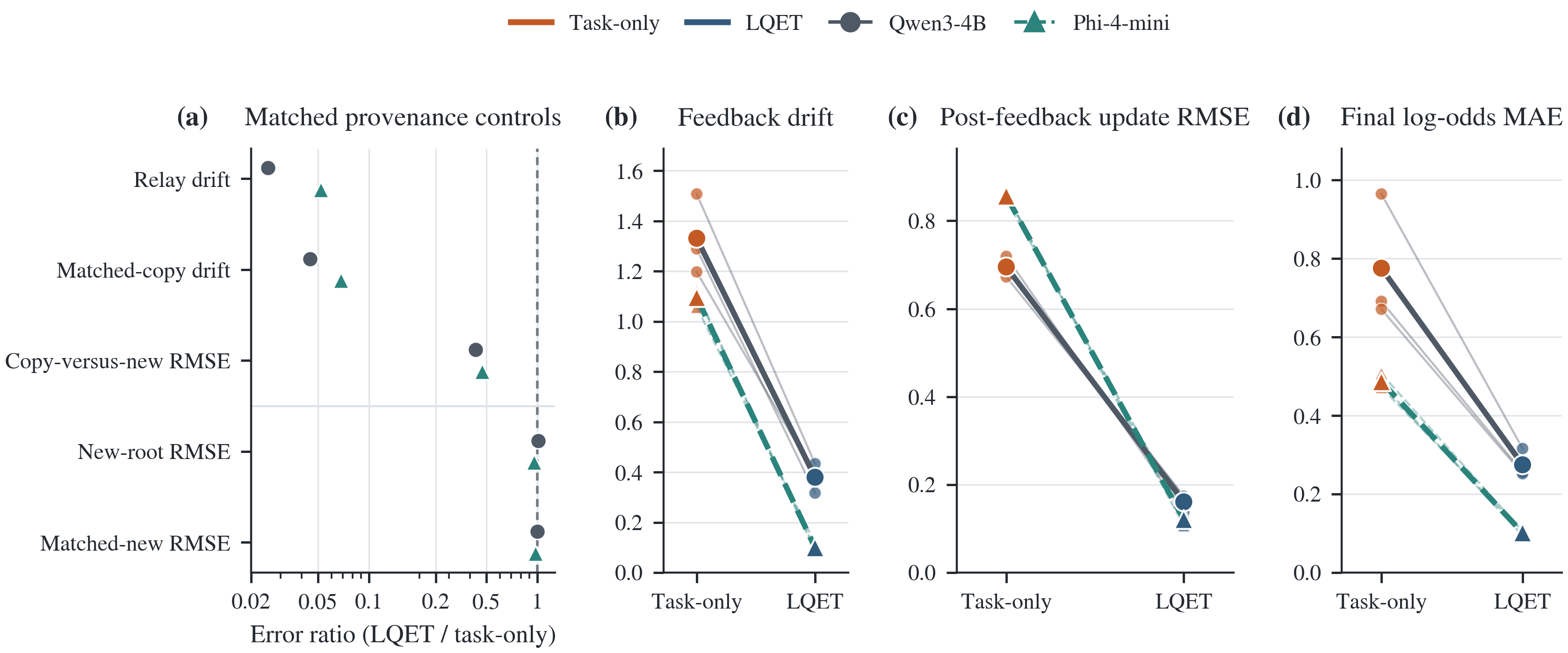}
    \caption{\textbf{Selective invariance and closed-loop feedback.} (a) \method{}/task-only error ratio on matched provenance controls, shown on a logarithmic x-axis; unity denotes no change, and marker and color identify the backbone. Inherited-report and copy-versus-new errors fall while new-evidence update errors stay near parity. (b-d) Thin segments pair matched adapters by seed; thick segments and larger markers are three-seed means. Endpoint color marks the method, while shape and line style mark the backbone. All panels are descriptive; lower is better.}
    \label{fig:selectivity-feedback}
\end{figure}

A model could lower relay drift by suppressing all updates. Matched copy-versus-new-evidence controls test this alternative directly (Figure~\ref{fig:selectivity-feedback}a). \method{} reduces base-to-relay and matched-copy drift by factors of approximately 15--40, while new-root and matched-new RMSE remain close to task-only. Here, new-root RMSE measures six supervised base-to-new-root updates, whereas the primary update RMSE measures the three relational update edges. In the tightly matched comparison, which holds the report author and displayed outcome fixed while varying supplied provenance, copy-versus-new RMSE drops from 1.0145 to 0.4383 on Qwen and from 0.9818 to 0.4630 on Phi. The corresponding zero-intercept copy-versus-new slopes rise from 0.027 to 0.782 and from 0.067 to 0.769. Within this supplied-lineage setting, the pattern is selective: the model becomes less sensitive to declared relays while retaining its response to declared new observations.

\subsection{Closed-loop feedback dynamics}
\label{sec:feedback}

We next place each trained model in a deterministic human-AI feedback stress test on a fixed 64-family subset of the recursive-relay split (Appendix~\ref{app:feedback}). Crossing each family with initial trust $\tau_0\in\{0.2,0.8\}$ yields 128 trajectories per trained model. In round $r$, score-derived confidence $c_r$, trust $\tau_r$, and the number of simulated human relays $m_r$ obey
\[
c_r=|2\sigma(z_r)-1|,\qquad
\tau_{r+1}=0.6\tau_r+0.4c_r,\qquad
m_r=1+\min\{2,\lfloor3\tau_{r+1}c_r\rfloor\}.
\]
Each of three rounds appends one AI-derived report and $m_r\in\{1,2,3\}$ simulated human relays that add no new evidence. A fresh root is then added to measure the subsequent update.

Feedback drift, post-feedback update RMSE, and final log-odds MAE improve in every backbone--seed pair (Figure~\ref{fig:selectivity-feedback}b-d). Mean reductions range from 64.6--76.9\% on Qwen and 79.1--90.9\% on Phi. This result connects the static provenance constraints to a dynamic failure mode: when confidence changes how many derived reports return, lower sensitivity to inherited reports also reduces feedback amplification.

\subsection{What each relation contributes}
\label{sec:ablation}

A Qwen seed-0 component study uses fresh, disjoint sets of 800 controlled-shift and 200 recursive-relay families. Changing the order coefficient from one to zero preserves all 13 forward views. Full \method{} improves the order-sensitivity gap (0.0426 to 0.0359), update RMSE (0.4105 to 0.4058), and supervised-view KL (0.0239 to 0.0232), while No-order has lower depth-8 drift (0.0753 versus 0.0959). The order term thus offers a measured trade-off, chiefly improving presentation-order consistency. Comparing No-order with Pair-invariance suggests that update supervision together with added composition views accounts for most of the compositional gain. Appendix~\ref{app:weight-sensitivity} details the weight comparisons and their interpretation.

\section{Related work}
\label{sec:related}

\paragraph{Belief updating and predictive consistency.}
Bayesian teaching can improve probabilistic reasoning in language models \citep{qiu2026bayesian}. Evaluations study responses to evidence reliability \citep{kim-etal-2025-evidence}, sequential belief trajectories \citep{samanta2026bayesbench}, and martingale conditions for in-context learning \citep{falck2024martingale}; Martingale Amortization aligns short- and longer-horizon predictions through self-consistency \citep{kim2026drift}. We instead ask whether supplied dependence changes the evidential force of otherwise matched reports.

\paragraph{Repetition bias and dependent social learning.}
Schuster et al. show that repetition can reverse language-model source preferences and reduce the effect by training on aligned repeated and unrepeated contexts \citep{schuster-etal-2026-whose}; this setup most closely overlaps our invariance component. Naphade finds that paraphrases can outweigh genuinely distinct support and documents presentation-order effects \citep{naphade-2026-rational}. Related work studies the chorus effect \citep{weaver2007chorus}, persuasion bias \citep{demarzo2003persuasion}, and correlation neglect \citep{enke2019correlation}. Graph-based methods correct data incest when topology and the probabilistic model are available \citep{brehard2007optimal,krishnamurthy2013misinformation}, while Jain and Krishnamurthy illustrate data incest among interacting language-model agents \citep{jain2025interacting}. Within our controlled setting, \method{} extends repetition invariance to explicit multi-hop lineage, exact reliability-conditioned targets, new-root updates across relay histories, composition, and presentation order.

\paragraph{Feedback loops and evidence propagation.}
Language-model outputs can enter later contexts and create failures that static evaluation misses \citep{pan2024feedback}; bidirectional alignment likewise emphasizes that model outputs shape later human inputs \citep{shen2025bidirectional}. Multi-agent evaluation traces evidence propagation through later messages \citep{yan2026distributed}. Our simulation isolates how score-dependent return of inherited reports interacts with provenance-sensitive updating.

\paragraph{Provenance, aggregation, and dependence discovery.}
Database provenance tracks information origins \citep{buneman2001why} and composes dependencies on input records \citep{green2007provenance}; natural-language claim-provenance graphs represent how claims originate and evolve \citep{zhang2020provenance}. GEAR aggregates evidence over graphs of retrieved sentences rather than report lineage \citep{zhou-etal-2019-gear}. Dong et al. infer copying and source dependence from output agreement \citep{dong2009source} and update histories \citep{dong2009dynamic}. \method{} assumes parent-report links are supplied and learns to map them to consistent belief updates, rather than retrieving evidence or discovering hidden dependence.

\section{Scope and limitations}
\label{sec:scope}

This study isolates provenance-aware updating in a controlled binary setting, enabling exact tests of evidence accounting across two 4B-class backbones and three paired seeds. Source-reliability uncertainty is modeled; supplied lineage is assumed correct. Transfer to larger models, naturalistic tasks, and incomplete or unreliable lineage remains to be tested (Appendix~\ref{app:lineage-scope}). The forced-choice \yes-\no{} score assesses binary preference; free-form probability estimation and downstream decision quality require separate evaluation.

Comparisons match training families and optimizer updates but differ in transcript-view exposure, so the reported gains characterize the complete training protocol. Under held-out prior and calibration supports, lower update RMSE accompanies higher supervised-view KL, identifying numerical generalization as a remaining challenge alongside improved relational consistency. The confidence-dependent simulation isolates report recirculation under a prescribed trust rule; studies with human participants are needed to assess how this mechanism operates in practical human-AI interaction.

\section{Conclusion}
\label{sec:conclusion}

\method{} formulates evidence accounting as a relational learning problem, connecting predictions on matched transcripts through provenance invariance and new-evidence update consistency. Matched controls distinguish the learned stability from indiscriminate update suppression, while the confidence-dependent feedback simulation demonstrates reduced amplification when model predictions shape future inputs. Supervision thus extends from targets for individual transcripts to consistent updates across relay histories. Provenance gives this learning target a concrete structure: which changes in context warrant changes in belief.

\bibliographystyle{plainnat}
\bibliography{references}

\clearpage
\appendix
\raggedbottom
\section{Exact Bayesian target and benchmark construction}
\label{app:target}

\subsection{Beta--Bernoulli derivation}

After conditioning on the public calibration record, fix $Y=y$ and a source $s$. Among the distinct relevant roots from $s$ in $E(T)$, let $C_s(y,T)$ count outcomes that agree with $y$ and $W_s(y,T)$ count outcomes that disagree. Integrating the source reliability in Equation~\ref{eq:reliability} gives

\begin{equation}
P\bigl((x_e)_{e\in E(T):\,s(e)=s}\mid Y=y,c_s,w_s\bigr)
=\frac{B(\alpha_s+C_s(y,T),\beta_s+W_s(y,T))}
       {B(\alpha_s,\beta_s)}.
\end{equation}
Multiplying over sources and taking posterior log-odds yields Equation~\ref{eq:bayes-target}. The terms $\log B(\alpha_s,\beta_s)$ cancel between the two values of $Y$, giving
\begin{equation}
\begin{aligned}
t(T)
&=\log\frac{\pi}{1-\pi}\\
&\quad+\sum_s\left[
    \log B(\alpha_s+n_{s,+}(T),\beta_s+n_{s,-}(T))
    -\log B(\alpha_s+n_{s,-}(T),\beta_s+n_{s,+}(T))
\right].
\end{aligned}
\label{eq:compact-target}
\end{equation}

Multiple distinct roots from one source are conditionally independent given $(Y,R_s)$ but become dependent after $R_s$ is marginalized. Descendant reports carrying the same root do not create repeated observations because $E(T)$ contains each root once.

The target is the exact Bayesian posterior under the benchmark's binary source-reliability model. The \texttt{matched-new} and \texttt{counter} views are constructed controls rather than standard sampled views: the former is matched to the copied-report control in author and displayed outcome but declares a new measurement, whereas the latter displays the opposing outcome. Their targets are computed from the evidence declared in each prompt, allowing provenance and outcome polarity to vary separately.

\subsection{Report provenance and validity}

Every family is validated as a directed acyclic report graph (DAG). Each evidence root has exactly one introducing report; any later report carrying that root includes an explicit link to a parent report that also carries it. A pure relay introduces no root. A mixed report can inherit one or more roots and introduce a new root at the same node. Roots marked irrelevant to the query are excluded from $E(T)$. These rules make evidence multiplicity a property of provenance rather than the number of textual mentions.

A provenance equivalence class contains transcripts with the same set $E(T)$ of relevant evidence roots. Holding the prior and public calibration records fixed, equality of source-by-outcome counts is sufficient for the same Bayesian target under this benchmark, but it does not define equivalence-class membership: distinct roots from the same source with the same outcome remain distinct evidence.

\section{Matched transcript views}
\label{app:views}

Table~\ref{tab:views} lists the 14 matched views. Latent root labels such as $e_0$, $e_a$, and $e_b$ are construction notation and are never rendered in a prompt. \method{} trains on 13 views; $c_{00}$ is the same rendered transcript as $h_{00}$ and contributes no distinct training term. The primary task-only and \method{} runs are evaluated on all 14 views in the static IID and controlled-shift sets.

\begin{table}[h]
\centering
\caption{\textbf{Matched transcript views.} \emph{Supervised} denotes a pointwise soft target; \emph{invariance} and \emph{update} denote relational loss terms.}
\label{tab:views}
\small
\setlength{\tabcolsep}{4.0pt}
\begin{tabularx}{\textwidth}{>{\raggedright\arraybackslash}p{3.1cm}>{\raggedright\arraybackslash}X>{\raggedright\arraybackslash}p{3.1cm}>{\raggedright\arraybackslash}p{3.4cm}}
\toprule
View & View construction & Bayesian-target relation & \method{} role \\
\midrule
$h_{00}$ & Base observation carrying root $e_0$ & base target & supervised; invariance/update reference \\
$h_{10}$ & Base plus a relay DAG; no new root & $t(h_{10})=t(h_{00})$ & invariance; update reference \\
$h_{01}$ & Base plus new root $e_h$ & new-evidence target & supervised; invariance/update target \\
$h_{11}$ & Relay DAG plus a mixed report introducing $e_h$ & $t(h_{11})=t(h_{01})$ & invariance; update endpoint \\
$c_{00}$ & Same rendered base as $h_{00}$ & $t(c_{00})=t(h_{00})$ & static evaluation only \\
$c_{10}$ & Base plus new root $e_a$ & new-evidence target & supervised; update reference \\
$c_{01}$ & Base plus new root $e_b$ & new-evidence target & supervised; update reference \\
$c_{11}$ & Base plus both $e_a$ and $e_b$ & updates add under source distinctness & update endpoint; order \\
$c_{11}^{\mathrm{rev}}$ & Same roots in reversed presentation order & $t(c_{11}^{\mathrm{rev}})=t(c_{11})$ & order \\
\texttt{matched-copy} & Matched author/outcome; carries the base root through an explicit parent-report link & $t=t(h_{00})$ & invariance \\
\texttt{matched-new} & Matched author/outcome; declares a new measurement and root & matched new-root target & supervised \\
\texttt{same-source} & New root from the base source & source-coupled target & supervised \\
\texttt{irrelevant} & Adds a root marked irrelevant to the query & $t=t(h_{00})$ & invariance \\
\texttt{counter} & Adds a controlled root with the opposing outcome & controlled new-root target & supervised \\
\bottomrule
\end{tabularx}
\end{table}

The base root $e_0$ and the two composition roots $e_a$ and $e_b$ come from three distinct sources. The core-block root $e_h$ is separate from these composition roots. Consequently,
\begin{equation}
t(c_{11})-t(c_{00})=
[t(c_{10})-t(c_{00})]+[t(c_{01})-t(c_{00})].
\end{equation}
The benchmark does not assume that arbitrary additions from the same source are additive.

\section{Training and evaluation details}
\label{app:protocol}

\subsection{Training views and objective variants}

Table~\ref{tab:training-controls} summarizes the training views
and objective variants.

\begin{table}[h]
\centering
\caption{\textbf{Training protocols.} Methods share training families and 500 optimizer updates.}
\label{tab:training-controls}
\small
\begin{tabular}{lcl}
\toprule
Method & Training views & Objective \\
\midrule
Task-only & 7 & $\mathcal L_{\mathrm{task}}$ \\
Pair-invariance & 11 & $\mathcal L_{\mathrm{task}}+4\mathcal L_{\mathrm{inv}}$ \\
No-order & 13 & $\mathcal L_{\mathrm{task}}+4\mathcal L_{\mathrm{inv}}+2\mathcal L_{\mathrm{update}}$ \\
Full \method{} & 13 & Equation~\ref{eq:lqet} \\
\bottomrule
\end{tabular}
\end{table}

\subsection{Optimization configuration}

\begin{table}[h]
\centering
\caption{\textbf{Shared training configuration.}}
\label{tab:hyperparameters}
\small
\setlength{\tabcolsep}{4.0pt}
\begin{tabularx}{\textwidth}{lXlX}
\toprule
Item & Value & Item & Value \\
\midrule
Training families & 4,000 & Validation families & 400 \\
Dataset seed & 17 & Training seeds & 0, 1, 2 \\
Optimizer updates & 500 & Epochs & 2 \\
Families/microbatch & 1 & Gradient accumulation & 16 \\
Optimizer & AdamW & Learning rate & $2\times10^{-4}$ \\
Weight decay & 0 & Warmup & 15 updates \\
Schedule & cosine & Gradient clip & 1.0 \\
Maximum length & 768 & Truncated prompts & 0 \\
Quantization & NF4 with double quantization & Compute dtype & BF16 \\
LoRA rank & 32 & LoRA $\alpha$ / dropout & 64 / 0.05 \\
LoRA targets & all linear modules & Huber $\delta$ & 0.25 \\
Loss weights & task/invariance/update/order: $1{:}4{:}2{:}1$ & Checkpoint & final update \\
\bottomrule
\end{tabularx}
\end{table}

The loss weights, Huber threshold, and optimization settings were fixed before the reported multi-seed Qwen runs and transferred unchanged to Phi; no validation-set hyperparameter search was conducted. Validation results served as run diagnostics only. The final checkpoint is evaluated without early stopping or validation-based checkpoint selection.

\subsection{Generator supports and prompt format}

Table~\ref{tab:generator-supports} specifies the discrete benchmark supports. Each controlled shift changes only the named factor group; all remaining factors use the training support.

\begin{table}[h]
\centering
\caption{\textbf{Benchmark-generator supports.} Calibration entries are numbers of correct and incorrect historical calibration trials, $(c,w)$.}
\label{tab:generator-supports}
\small
\setlength{\tabcolsep}{3.5pt}
\begin{tabularx}{\textwidth}{>{\raggedright\arraybackslash}p{2.8cm}>{\raggedright\arraybackslash}X>{\raggedright\arraybackslash}X}
\toprule
Factor & Training, validation, and IID support & Controlled held-out support \\
\midrule
Domain and names & Four domains: beacon status, valve position, package authenticity, and server security; 26 source names & Three domains: soil contamination, sample radioactivity, and component magnetization; 26 disjoint source names \\
Report graph & Chains of depth 1--3; stars with 1--8 relays & Depth shift: chains of depth 4--8. Topology shift: diamonds with 3 or 4 relays; fan-in graphs with branch depth 1 or 2 (3 or 5 relays total) \\
Rendering template & Direct and contrast renderings & Record-style and basis-style renderings \\
Prior $\pi$ & $\{0.20,\allowbreak 0.35,\allowbreak 0.50,\allowbreak 0.65,\allowbreak 0.80\}$ & $\{0.125,\allowbreak 0.275,\allowbreak 0.425,\allowbreak 0.575,\allowbreak 0.725,\allowbreak 0.875\}$ \\
Calibration $(c,w)$ & $c\in\{4,\ldots,9\}$ and $w\in\{1,2,3\}$ & $\{(0,0),\allowbreak (0,3),\allowbreak (1,4),\allowbreak (2,0),\allowbreak (2,5),\allowbreak (4,4),\allowbreak (4,10),\allowbreak (10,0),\allowbreak (10,4),\allowbreak (10,10),\allowbreak (11,0),\allowbreak (11,5)\}$ \\
\bottomrule
\end{tabularx}
\end{table}

Each family has four calibrated evidence sources; additional names can appear only as relay authors and contribute no independent observation. The prior and each calibrated source's record are sampled uniformly from the applicable supports. For generatively sampled evidence roots, the generator draws $R_s\sim\operatorname{Beta}(c_s+1,w_s+1)$, samples $Y$ with $P(Y=+1)=\pi$, and draws an outcome that agrees with $Y$ with probability $R_s$. Separate SHA-256-derived random-number streams govern evidence, calibration, domain, graph, and surface choices. A fixed SHA-256 rule assigns candidate family seeds to training, validation, and test partitions in an 80/10/10 ratio. Automated checks reject any cross-partition overlap in family identifiers, complete benchmark states, latent evidence configurations, or rendered prompts.

\noindent
The benchmark intentionally supplies report lineage. In this representative prompt from the training distribution, parent citations and report text identify an independently acquired observation and its relay; the evaluated task is to use these declarations consistently across matched transcripts and report graphs.
\begin{quote}
\small\raggedright
\textbf{Question:} Is the valve open?\\
\textbf{Prior probability of YES:} 0.650\\
\textbf{Calibration records:}\\
-- Ava previously made 8 correct and 2 incorrect measurements.\\
-- Ben previously made 5 correct and 1 incorrect measurements.\\
-- Dara previously made 6 correct and 3 incorrect measurements.\\
-- Emil previously made 4 correct and 2 incorrect measurements.\\
\textbf{Reports:}\\
1. [R1] Ava: Cites: none. Ava independently ran a new position check without seeing any other report. The recorded outcome was OPEN.\\
2. [R2] Cleo: Cites: R1. Cleo read the cited predecessor report, ran no new measurement, and recorded the inherited outcome OPEN.\\
\textbf{Instruction:} Estimate the posterior probability that the answer is YES.
\end{quote}
The displayed instruction is the verbatim instruction used in the experiments. We do not decode or parse a generated numeric answer. Instead, the reported quantity is the forced-choice \yes{} score in Equation~\ref{eq:model-score}, conditional on the next token belonging to $\{\yes,\no\}$. For each static IID or OOD view, the rendered prompt is identical across methods.

\subsection{Static and recursive-relay evaluation sets}

Each run is evaluated on a static 400-family IID set and five static controlled-shift sets of 160 families each:
\begin{itemize}[leftmargin=*,itemsep=1pt,topsep=2pt]
    \item \textbf{domain/names}: held-out domains and source names;
    \item \textbf{depth}: relay depths 4--8;
    \item \textbf{topology}: diamond and fan-in report graphs;
    \item \textbf{template}: held-out rendering templates; and
    \item \textbf{prior/calibration}: held-out supports of priors and calibration counts.
\end{itemize}
All 14 matched views are scored on these static sets. Within each split, the ordered family roster is shared across methods, training seeds, and backbones.

A separate set of 200 recursive-relay families supports the depth and closed-loop feedback evaluations. For each family, one provenance-preserving chain is extended through depth 8; depths 1, 2, 4, and 8 are nested snapshots of that chain. The new-root probe at each snapshot is evaluated on a separate branch and is not included in later relay steps. The fixed ordered list of recursive-relay families, lineage schedule, and rendering rule is shared across methods, seeds, and backbones. Realized prompt text becomes model-specific after the first prediction because each appended relay quotes that model's preceding forced-choice score. Training, validation, static evaluation, and recursive-relay partitions are mutually disjoint in benchmark instances, evidence roots, family identifiers, and rendered prompts. Every scored prompt fits within the length limit without truncation.

\subsection{Endpoint definitions and bootstrap}

For a fixed backbone and method, let seed $j\in\{0,1,2\}$ and suppress the backbone and method indices. Let
$\mathcal E_{\mathrm{upd}}=\{(h_{10},h_{11}),(c_{10},c_{11}),(c_{01},c_{11})\}$
contain the three oriented new-evidence update edges from Equation~\ref{eq:update}, written $(v^-,v^+)$. For family $i$ and view $v$, write $z_{j,i}(v)=z_{\theta_j}(T_{i,v})$, $t_i(v)=t(T_{i,v})$, $\targetprob_i(v)=\sigma(t_i(v))$, and $\hat p_{j,i}(v)=\sigma(z_{j,i}(v))$. On the 800 pooled controlled-shift families, the new-evidence update RMSE is
\begin{equation}
\operatorname{RMSE}^{\mathrm{update}}_j
=\sqrt{\frac{1}{800\cdot3}\sum_{i=1}^{800}
\sum_{(v^-,v^+)\in\mathcal E_{\mathrm{upd}}}
\left([z_{j,i}(v^+)-z_{j,i}(v^-)]-[t_i(v^+)-t_i(v^-)]\right)^2}.
\end{equation}

For recursive-relay family $i$, let $z^{\mathrm{base}}_{j,i}$ be the base score and $z^{\mathrm{relay},8}_{j,i}$ the score after eight provenance-preserving relays, before any new root is added. The seed-level depth-8 drift is
\begin{equation}
D^{\mathrm{relay}}_{8,j}
=\frac{1}{200}\sum_{i=1}^{200}
\left|z^{\mathrm{relay},8}_{j,i}-z^{\mathrm{base}}_{j,i}\right|.
\end{equation}

Let
\[
\mathcal A=\{h_{00},h_{01},c_{10},c_{01},\text{same-source},\text{counter},\text{matched-new}\}
\]
be the seven supervised views. Their seed-level supervised-view KL is
\begin{equation}
K^{\mathrm{sup}}_j
=\frac{1}{800|\mathcal A|}\sum_{i=1}^{800}\sum_{v\in\mathcal A}
\KL\!\left(\operatorname{Bern}(\targetprob_i(v))\,\middle\|\,
\operatorname{Bern}(\hat p_{j,i}(v))\right).
\end{equation}
For any seed-level endpoint $q_j$ above, the reported backbone point estimate is $\bar q=\frac13\sum_{j=0}^{2}q_j$. In particular, each RMSE is square-rooted within its seed before the three seed-level values are averaged.

Because every trained model is evaluated on the same family rosters, training seeds and evaluation families form a crossed rather than nested design. Using random seed 20260829, each of 10,000 bootstrap replicates independently draws three paired seed labels with replacement and one common with-replacement family sample. For pooled controlled-shift endpoints, the family draw is stratified by resampling 160 families within each of the five fixed shifts; IID endpoints use one 400-family draw, and depth-8 drift uses one 200-family recursive-relay draw. The sampled family indices are reused across every selected seed occurrence and both methods, and all views or relay snapshots from a family remain grouped. We recompute each method-specific endpoint within seed, including taking the square root for RMSE, average over the three sampled seed occurrences, and form the \method{}-minus-task-only contrast. We report the 2.5th and 97.5th percentiles without multiplicity adjustment.

\section{Primary results by seed and evaluation regime}
\label{app:primary-details}

\subsection{Every training seed}

\begin{table}[H]
\centering
\caption{\textbf{Per-seed primary endpoints.} Effects are \method{} minus task-only and are computed from unrounded estimates; lower is better.}
\label{tab:per-seed-primary}
\small
\setlength{\tabcolsep}{4.5pt}
\begin{tabular}{llcrrr}
\toprule
Backbone & Endpoint & Seed & Task-only & \method{} & Effect \\
\midrule
Qwen3-4B & Update RMSE & 0 & 0.6596 & 0.3901 & $-0.2695$ \\
& & 1 & 0.6895 & 0.4143 & $-0.2752$ \\
& & 2 & 0.6958 & 0.3971 & $-0.2987$ \\
& Depth-8 drift & 0 & 1.2172 & 0.0931 & $-1.1241$ \\
& & 1 & 0.8875 & 0.1078 & $-0.7797$ \\
& & 2 & 0.9319 & 0.0616 & $-0.8703$ \\
& Supervised-view KL & 0 & 0.0212 & 0.0226 & $+0.0014$ \\
& & 1 & 0.0227 & 0.0258 & $+0.0031$ \\
& & 2 & 0.0205 & 0.0229 & $+0.0024$ \\
\addlinespace
Phi-4-mini & Update RMSE & 0 & 0.7464 & 0.4074 & $-0.3391$ \\
& & 1 & 0.7439 & 0.4199 & $-0.3240$ \\
& & 2 & 0.7479 & 0.4160 & $-0.3319$ \\
& Depth-8 drift & 0 & 1.0375 & 0.1156 & $-0.9219$ \\
& & 1 & 1.1775 & 0.1313 & $-1.0463$ \\
& & 2 & 1.1088 & 0.1063 & $-1.0025$ \\
& Supervised-view KL & 0 & 0.0279 & 0.0281 & $+0.0001$ \\
& & 1 & 0.0277 & 0.0296 & $+0.0019$ \\
& & 2 & 0.0285 & 0.0301 & $+0.0015$ \\
\bottomrule
\end{tabular}
\end{table}

\subsection{IID test results}

\begin{table}[H]
\centering
\caption{\textbf{Results on the 400-family IID test set.} Point estimates average three paired training seeds. Effects are \method{} minus task-only; intervals are 95\% paired crossed seed-by-family bootstrap intervals. Lower is better.}
\label{tab:iid-results}
\small
\setlength{\tabcolsep}{4.2pt}
\begin{tabular}{llrrr}
\toprule
Backbone & Endpoint & Task-only & \method{} & Effect [95\% interval] \\
\midrule
Qwen3-4B & Update RMSE & 0.5635 & \textbf{0.0515} & $-0.5119$ [$-0.5395,-0.4834$] \\
& Supervised-view KL & 0.00052 & \textbf{0.00023} & $-0.00029$ [$-0.00036,-0.00022$] \\
\addlinespace
Phi-4-mini & Update RMSE & 0.6354 & \textbf{0.0865} & $-0.5489$ [$-0.5772,-0.5200$] \\
& Supervised-view KL & 0.00156 & \textbf{0.00061} & $-0.00094$ [$-0.00104,-0.00084$] \\
\bottomrule
\end{tabular}
\end{table}

\subsection{Controlled-shift decomposition}

\begin{table}[H]
\centering
\caption{\textbf{Results by controlled shift.} Descriptive three-seed means are shown as task-only $\rightarrow$ \method{}. Update RMSE is measured in log-odds units and KL in nats. Each shift has 160 families per seed; lower is better.}
\label{tab:ood-axes}
\small
\setlength{\tabcolsep}{4.6pt}
\begin{tabular}{llcc}
\toprule
Backbone & Shift & Update RMSE & Supervised-view KL \\
\midrule
Qwen3-4B & Domain/names & $0.5717\rightarrow0.1770$ & $0.00673\rightarrow0.00546$ \\
& Depth & $0.5371\rightarrow0.0511$ & $0.00052\rightarrow0.00026$ \\
& Topology & $0.6287\rightarrow0.0500$ & $0.00055\rightarrow0.00025$ \\
& Template & $0.5903\rightarrow0.0928$ & $0.00076\rightarrow0.00080$ \\
& Prior/calibration & $0.9817\rightarrow0.8694$ & $0.09877\rightarrow0.11196$ \\
\addlinespace
Phi-4-mini & Domain/names & $0.6278\rightarrow0.1129$ & $0.00685\rightarrow0.00202$ \\
& Depth & $0.6146\rightarrow0.0839$ & $0.00151\rightarrow0.00061$ \\
& Topology & $0.7009\rightarrow0.0842$ & $0.00151\rightarrow0.00062$ \\
& Template & $0.6556\rightarrow0.1445$ & $0.00180\rightarrow0.00131$ \\
& Prior/calibration & $1.0439\rightarrow0.9005$ & $0.12853\rightarrow0.14164$ \\
\bottomrule
\end{tabular}
\end{table}

The prior-and-calibration shift, which requires generalization to held-out numerical supports for both the class prior and source-calibration records, is the most difficult shift for both methods. \method{} still lowers update RMSE from 0.9817 to 0.8694 on Qwen (11.4\%) and from 1.0439 to 0.9005 on Phi (13.7\%), although supervised-view KL rises by approximately 0.013 nats on each backbone. Across the other four shifts, supervised-view KL decreases in seven of eight backbone--shift comparisons and changes by only $4\times10^{-5}$ in the remaining comparison.

\subsection{Recursive-relay depth trajectory}

\begin{table}[H]
\centering
\caption{\textbf{Drift across recursive-relay depth.} Values are mean absolute pre-update log-odds drift over three seeds and correspond to Figure~\ref{fig:controlled-generalization}(a,b).}
\label{tab:depth-values}
\small
\begin{tabular}{llrrrr}
\toprule
Backbone & Method & Depth 1 & Depth 2 & Depth 4 & Depth 8 \\
\midrule
Qwen3-4B & Task-only & 0.3911 & 0.7548 & 0.9932 & 1.0122 \\
& \method{} & \textbf{0.0385} & \textbf{0.0469} & \textbf{0.0594} & \textbf{0.0875} \\
Phi-4-mini & Task-only & 0.2529 & 0.5454 & 0.7879 & 1.1079 \\
& \method{} & \textbf{0.0629} & \textbf{0.0619} & \textbf{0.0842} & \textbf{0.1177} \\
\bottomrule
\end{tabular}
\end{table}

\section{Prompt sufficiency and selective-updating controls}
\label{app:identifiability}

\subsection{Bayesian-target recovery from rendered prompts}

A deterministic rule-based calculator receives only the rendered prompt. It parses the prior, calibration records, report text, relevance flags, and explicit parent-report links; it does not read Bayesian targets, hidden root identifiers, or family metadata. It successfully processes all $1{,}200\times14=16{,}800$ transcript-view instances in the static IID and controlled-shift sets. Its maximum absolute difference from the exact Bayesian log-odds computed directly from the benchmark state is below $5\times10^{-15}$, confirming that the Bayesian target is uniquely determined by the rendered prompt in this benchmark.

\subsection{Matched copy-versus-new-root controls}

Let $I=800$ denote the pooled controlled-shift families for training seed $j$, and write $z_{j,i}(v)$ and $t_i(v)$ for the model and exact Bayesian log-odds of view $v$. For the six supervised new-root views
\[
\mathcal V_{\mathrm{new}}=\{h_{01},c_{10},c_{01},\text{same-source},\text{counter},\text{matched-new}\},
\]
define
\begin{equation}
r^{\mathrm{new}}_{j,i}(v)=
\bigl[z_{j,i}(v)-z_{j,i}(h_{00})\bigr]
-\bigl[t_i(v)-t_i(h_{00})\bigr].
\end{equation}
The relay and matched-copy metrics are the mean absolute changes along their provenance-preserving edges,
\begin{align}
D_{\mathrm{relay},j}
&=\frac{1}{I}\sum_i\left|z_{j,i}(h_{10})-z_{j,i}(h_{00})\right|,\\
D_{\mathrm{copy},j}
&=\frac{1}{I}\sum_i\left|z_{j,i}(\text{matched-copy})-z_{j,i}(h_{00})\right|.
\end{align}
The new-root RMSE averages all six supervised new-root views, whereas matched-new RMSE uses only the \texttt{matched-new} control:
\begin{align}
\operatorname{RMSE}_{\mathrm{new},j}
&=\sqrt{\frac{1}{6I}\sum_i\sum_{v\in\mathcal V_{\mathrm{new}}}r^{\mathrm{new}}_{j,i}(v)^2},\\
\operatorname{RMSE}_{\mathrm{matched\text{-}new},j}
&=\sqrt{\frac{1}{I}\sum_i r^{\mathrm{new}}_{j,i}(\text{matched-new})^2}.
\end{align}
Finally, the copy-versus-new contrast holds report author and displayed outcome fixed while changing whether the report carries the base root through an explicit parent-report link or declares a new measurement. With
\begin{align}
\widehat\delta_{j,i}
&=z_{j,i}(\text{matched-new})-z_{j,i}(\text{matched-copy}),\\
\delta_i^*
&=t_i(\text{matched-new})-t_i(\text{matched-copy}),
\end{align}
we report
\begin{align}
\operatorname{RMSE}_{\mathrm{copy\text{-}new},j}
&=\sqrt{\frac{1}{I}\sum_i(\widehat\delta_{j,i}-\delta_i^*)^2},\\
\operatorname{Slope}_j
&=\frac{\sum_i\delta_i^*\widehat\delta_{j,i}}
{\sum_i(\delta_i^*)^2}.
\end{align}
The slope is the zero-intercept regression coefficient of the predicted copy-versus-new change on the exact Bayesian change; its ideal value is one. Table~\ref{tab:selective-full} averages each seed-level metric over the three training seeds.

\begin{table}[h]
\centering
\caption{\textbf{Provenance-selectivity metrics.} Values are descriptive means over three seeds. Lower is better except for the copy-versus-new slope, whose ideal value is one.}
\label{tab:selective-full}
\small
\setlength{\tabcolsep}{2.8pt}
\begin{tabular}{llrrrrrr}
\toprule
Backbone & Method & \shortstack{Relay\\drift} & \shortstack{Copy\\drift} & \shortstack{New-root\\RMSE} & \shortstack{Matched-new\\RMSE} & \shortstack{Copy-vs-new\\RMSE} & \shortstack{Copy-vs-new\\slope} \\
\midrule
Qwen3-4B & Task-only & 1.0463 & 0.8338 & \textbf{0.3980} & \textbf{0.4326} & 1.0145 & 0.027 \\
& \method{} & \textbf{0.0264} & \textbf{0.0373} & 0.4043 & 0.4328 & \textbf{0.4383} & \textbf{0.782} \\
Phi-4-mini & Task-only & 1.0377 & 0.7813 & 0.4352 & 0.4682 & 0.9818 & 0.067 \\
& \method{} & \textbf{0.0540} & \textbf{0.0533} & \textbf{0.4185} & \textbf{0.4582} & \textbf{0.4630} & \textbf{0.769} \\
\bottomrule
\end{tabular}
\end{table}

The \texttt{matched-copy} and \texttt{matched-new} views share the report author and displayed outcome; their supplied provenance differs because one relays the base observation and the other declares a new measurement. Because provenance status is intentionally explicit, this control tests whether the model uses supplied provenance while holding author and outcome fixed. Together with retained new-root update accuracy, the result argues against reliance on author/outcome alone and against uniform suppression of log-odds changes.

\section{Closed-loop feedback details}
\label{app:feedback}

\begin{figure}[t]
  \centering
  \includegraphics[width=\linewidth]{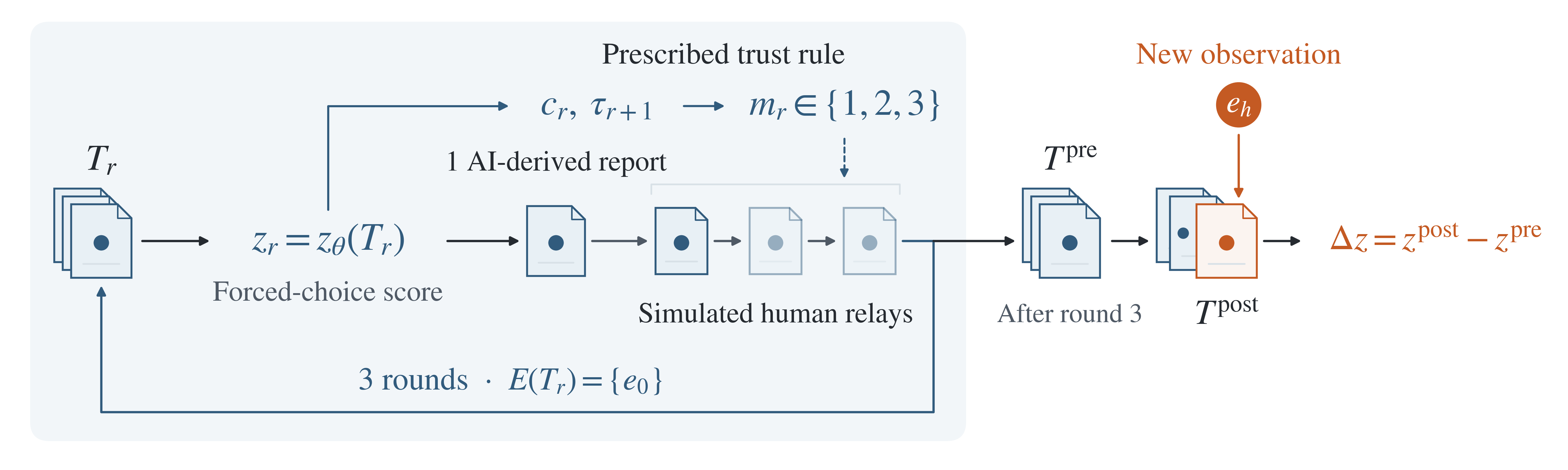}
  \caption{\textbf{Confidence-dependent feedback simulation.}
  A prescribed trust rule controls how many simulated human relays follow each
  AI-derived report. After three rounds without new evidence, a fresh observation
  is added. The resulting log-odds change is compared with the exact Bayesian update.}
  \label{fig:feedback-mechanism}
\end{figure}

The closed-loop feedback analysis (Figure~\ref{fig:feedback-mechanism}) uses the first 64 families in the fixed ordered list of the same 200 recursive-relay families described in Section~\ref{app:protocol}. The subset rule and all simulator constants were fixed before evaluation. The two trust arms span low- and high-trust initial conditions, while three rounds with at most three relays per round define a bounded stress test; the same deterministic family prefix and settings preserve paired comparisons across methods, seeds, and backbones. Each family is evaluated at initial trust values 0.2 and 0.8, producing 128 trajectories per model and seed. If $z_r$ is the current forced-choice log-odds, round $r$ applies
\begin{align}
p_r&=\sigma(z_r), & c_r&=|2p_r-1|,\\
\tau_{r+1}&=0.6\tau_r+0.4c_r,
&m_r&=1+\min\{2,\lfloor3\tau_{r+1}c_r\rfloor\}.
\end{align}
Here $c_r$ is confidence derived from the forced-choice score. The simulator appends one AI-derived report followed by $m_r\in\{1,2,3\}$ simulated human relays, then recomputes the model score from the rendered prompt. Each appended report includes an explicit link to its immediate parent report and introduces no new evidence root. After three rounds, the simulator appends an observation using the same new-root construction as $h_{01}$.

Let $z_i^0$ be the base log-odds, $z_{ia}^{\mathrm{pre}}$ the log-odds after three feedback rounds for family $i$ and trust condition $a$, $z_{ia}^{\mathrm{post}}$ the log-odds after the new-root observation, and $t_{ia}^{\mathrm{pre}},t_{ia}^{\mathrm{post}}$ the corresponding exact Bayesian log-odds. With $M=128$ trajectories, the endpoints are
\begin{align}
D_{\mathrm{feedback}}&=\frac{1}{M}\sum_{i,a}|z_{ia}^{\mathrm{pre}}-z_i^0|,\\
\operatorname{RMSE}_{\mathrm{postfb}}&=
\sqrt{\frac{1}{M}\sum_{i,a}
\left[(z_{ia}^{\mathrm{post}}-z_{ia}^{\mathrm{pre}})-(t_{ia}^{\mathrm{post}}-t_{ia}^{\mathrm{pre}})\right]^2},\\
\operatorname{MAE}_{\mathrm{final}}&=\frac{1}{M}\sum_{i,a}|z_{ia}^{\mathrm{post}}-t_{ia}^{\mathrm{post}}|.
\end{align}

\begin{table}[H]
\centering
\caption{\textbf{Closed-loop feedback results by training seed.} Lower is better.}
\label{tab:feedback-seeds}
\small
\setlength{\tabcolsep}{3.5pt}
\begin{tabular}{llcrrr}
\toprule
Backbone & Method & Seed & \shortstack{Feedback\\drift} & \shortstack{Post-feedback\\update RMSE} & \shortstack{Final log-odds\\MAE} \\
\midrule
Qwen3-4B & Task-only & 0 & 1.5078 & 0.6936 & 0.9644 \\
& \method{} & 0 & \textbf{0.4336} & \textbf{0.1732} & \textbf{0.3160} \\
& Task-only & 1 & 1.1973 & 0.6725 & 0.6909 \\
& \method{} & 1 & \textbf{0.3887} & \textbf{0.1733} & \textbf{0.2560} \\
& Task-only & 2 & 1.2891 & 0.7191 & 0.6708 \\
& \method{} & 2 & \textbf{0.3169} & \textbf{0.1362} & \textbf{0.2520} \\
\addlinespace
Phi-4-mini & Task-only & 0 & 1.0566 & 0.8477 & 0.4717 \\
& \method{} & 0 & \textbf{0.1172} & \textbf{0.1361} & \textbf{0.1041} \\
& Task-only & 1 & 1.1191 & 0.8601 & 0.5065 \\
& \method{} & 1 & \textbf{0.0938} & \textbf{0.1197} & \textbf{0.1056} \\
& Task-only & 2 & 1.1113 & 0.8610 & 0.4810 \\
& \method{} & 2 & \textbf{0.0869} & \textbf{0.1068} & \textbf{0.0947} \\
\bottomrule
\end{tabular}
\end{table}

\method{} also reduces the mean cumulative number of simulated human relays from 5.4349 to 4.4557 on Qwen and from 5.0469 to 4.3438 on Phi. Because the number of relays depends on score-derived confidence, these trajectories capture both reduced sensitivity to provenance-preserving reports and the associated reduction in how many such reports return.

\section{Component analysis}
\label{app:ablation}

The Qwen3-4B seed-0 component study uses fresh, disjoint evaluation sets: 800 static controlled-shift families (160 per shift) for update RMSE, supervised-view KL, and order-sensitivity gap, and 200 recursive-relay families for depth-8 drift. For the static controlled-shift families, define
\begin{equation}
G_{\mathrm{order}}=\frac{1}{800}\sum_{i=1}^{800}
\left|z_i(c_{11})-z_i(c_{11}^{\mathrm{rev}})\right|.
\end{equation}
Full \method{} and No-order both process 13 training views and use identical forward computation. No-order sets the order-loss weight to zero, so $c_{11}^{\mathrm{rev}}$ has no loss contribution. Pair-invariance processes 11 training views. Intervals use 10,000 paired family-bootstrap draws over the relevant fresh static or recursive-relay families.

\begin{table}[H]
\centering
\caption{\textbf{Qwen3-4B seed-0 component results.} Lower is better.}
\label{tab:no-order-values}
\small
\begin{tabular}{lrrrr}
\toprule
Method & Update RMSE & Depth-8 drift & Supervised-view KL & Order gap \\
\midrule
Pair-invariance & 0.5860 & 0.0984 & \textbf{0.0212} & 0.1184 \\
No-order & 0.4105 & \textbf{0.0753} & 0.0239 & 0.0426 \\
Full \method{} & \textbf{0.4058} & 0.0959 & 0.0232 & \textbf{0.0359} \\
\bottomrule
\end{tabular}
\end{table}

\begin{table}[H]
\centering
\caption{\textbf{Paired component effects.} Effects are No-order minus full \method{} and use unrounded estimates. Brackets are unadjusted 95\% paired family-bootstrap intervals; positive effects favor full \method{} for these lower-is-better endpoints.}
\label{tab:no-order-effects}
\small
\begin{tabular}{lrr}
\toprule
Endpoint & Effect & 95\% CI \\
\midrule
Update RMSE & $+0.00475$ & [$+0.00157,+0.00782$] \\
Depth-8 drift & $-0.02063$ & [$-0.03125,-0.01031$] \\
Supervised-view KL & $+0.00076$ & [$+0.00031,+0.00125$] \\
Order gap & $+0.00664$ & [$+0.00289,+0.01023$] \\
\bottomrule
\end{tabular}
\end{table}

\begin{figure}[H]
    \centering
    \includegraphics[width=\linewidth]{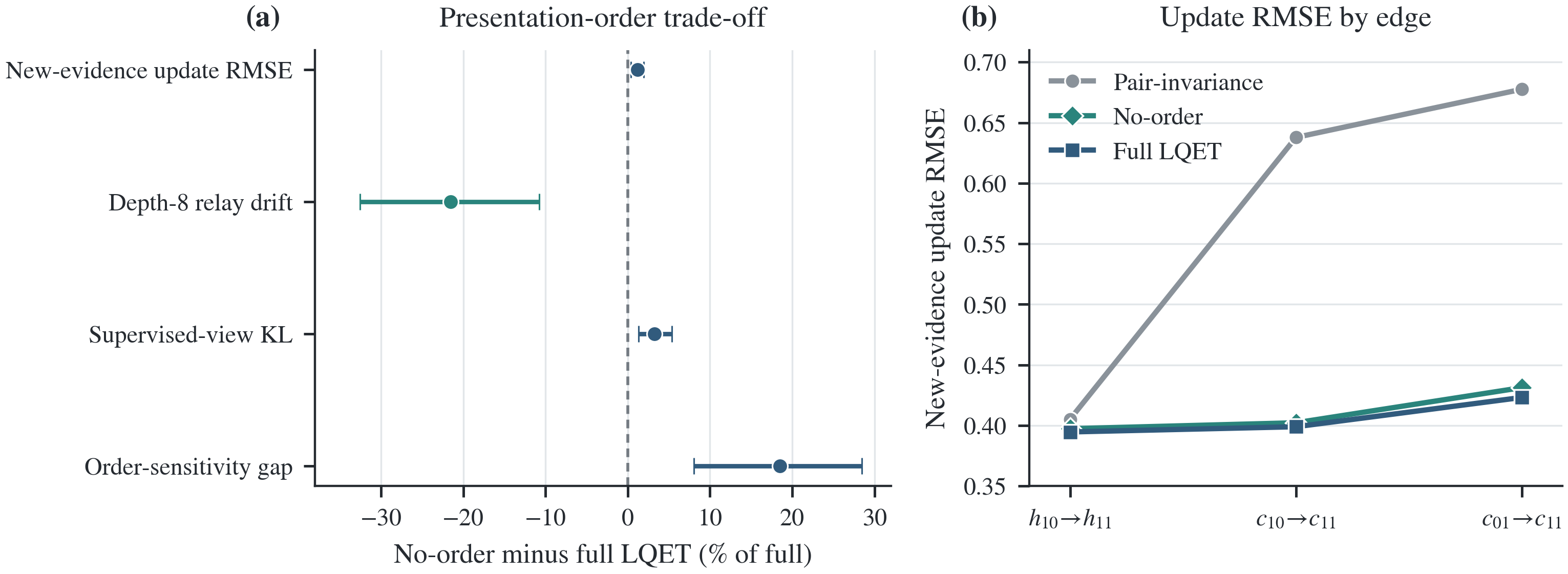}
    \caption{\textbf{Single-seed component trade-off.} (a) No-order-minus-full effects normalized by full \method{}, with 95\% paired family-bootstrap intervals; positive values favor full \method{} for the displayed lower-is-better endpoints. (b) Edgewise update RMSE across component variants.}
    \label{fig:ablation-tradeoff}
\end{figure}

The order term yields its clearest improvement on the order-sensitivity gap and modest improvements in update RMSE and supervised-view KL, while No-order has lower depth-8 drift for this seed (Figure~\ref{fig:ablation-tradeoff}a). Relative to Pair-invariance, No-order reduces update RMSE by 0.1755 and the order-sensitivity gap by 0.0759; the two composition edges account for most of the update improvement (Figure~\ref{fig:ablation-tradeoff}b). Because adding update consistency also adds the composition views, this comparison estimates their combined contribution rather than isolating the loss term alone.

\section{Provenance assumptions and objective weights}
\label{app:additional-analysis}

\subsection{Supplied lineage and scope of transfer}
\label{app:lineage-scope}

The evidence-accounting distinction also applies conceptually to a retrieved passage and faithful summaries or quotations that add no new observation. Recorded derivation would identify their shared origin; an additional observation would be assessed using its source reliability. Evaluating this behavior with naturalistic content remains a separate empirical test.

Source-reliability uncertainty concerns whether a distinct observation reports the state correctly; lineage uncertainty concerns which observations are distinct. The Bayesian target integrates the former and conditions on the latter. Misclassifying a relay as a new observation can count evidence twice, while misclassifying a new observation as a relay can discard evidence. Missing lineage can leave multiple evidence sets compatible with the visible reports. Furthermore, parent links and textual new-observation/relay declarations jointly encode provenance here: deleting a link may leave the same information recoverable from the text. A corruption evaluation would therefore need to distinguish missing cues, coherently misleading lineage, and contradictory graph/text cues, with a specified rule for interpreting each. The present experiments do not measure degradation under these corruptions.

\subsection{Loss scaling and coefficient sensitivity}
\label{app:weight-sensitivity}

The coefficients in Equation~\ref{eq:lqet} multiply component means. Within a family, the four invariance residuals each receive weight $4/4=1$, the three update residuals each receive weight $2/3$, and the order residual receives weight $1$. Each of the seven pointwise losses receives weight $1/7$. These fixed weights specify relative emphasis across the averaged losses; the cross-entropy and Huber gradients also depend on the current prediction errors. The Huber threshold $\delta=0.25$ is measured in log-odds residual units: errors below the threshold incur a quadratic penalty, while the derivative magnitude with respect to the residual is capped at $0.25$ beyond it. This limits the influence of large relational residuals without changing their zero-error target.

\paragraph{Compatible ideal targets.}
For nonnegative relational coefficients, consider
\[
\mathcal L_{\lambda}=\mathcal L_{\mathrm{task}}
+\lambda_{\mathrm{inv}}\mathcal L_{\mathrm{inv}}
+\lambda_{\mathrm{upd}}\mathcal L_{\mathrm{update}}
+\lambda_{\mathrm{ord}}\mathcal L_{\mathrm{order}}.
\]
In unrestricted score space, $z_i(v)=t_i(v)$ is a global minimizer for every such choice. To see this, subtract the mean target Bernoulli entropy over the supervised views from $\mathcal L_{\lambda}$: the remainder is the mean supervised-view KL plus nonnegative Huber penalties. The exact Bayesian scores make all these terms zero, using the stated source distinctness for the composition relations. Thus reweighting preserves a common ideal target; it can still change approximation and optimization error in a finite-capacity model trained for 500 updates. This analytical compatibility does not imply empirical insensitivity to the weights.

\paragraph{Available empirical evidence.}
The same configuration transfers from Qwen to Phi and improves the primary relational endpoints in all six backbone--seed pairs. The component study in Appendix~\ref{app:ablation} additionally compares order weights one and zero with identical 13-view forward computation. The Pair-invariance comparison jointly changes update supervision and composition-view exposure. These results establish fixed-configuration transfer and specific component trade-offs; no sweep of nonzero weights, Huber thresholds, or optimization settings is reported.

\end{document}